\documentclass[11pt]{article}
\usepackage{acl2016}
\usepackage{times}
\usepackage{listings}
\usepackage{latexsym}
\usepackage{graphicx}
\usepackage{comment,bm}
\usepackage{color,amsmath}
\usepackage[multiple]{footmisc}

\definecolor{gray}{rgb}{0.4,0.4,0.4}
\definecolor{darkblue}{rgb}{0.0,0.0,0.6}
\definecolor{cyan}{rgb}{0.0,0.6,0.6}
\usepackage[margin=1in,letterpaper]{geometry}
\lstdefinelanguage{XML}
{
  morestring=[b]",
  morestring=[s]{>}{<},
  morecomment=[s]{<?}{?>},
  stringstyle=\color{black},
  identifierstyle=\color{darkblue},
  keywordstyle=\color{cyan},
  morekeywords={xmlns,version,type}
}
\aclfinalcopy 

\title{Supervised and Unsupervised Ensembling for Knowledge Base Population}

\author{Nazneen Fatema Rajani\\
	   Dept. Of Computer Science\\
	    University of Texas at Austin\\
	    {\tt nrajani@cs.utexas.edu}
	  \And
	Raymond J. Mooney\\
  	 Dept. Of Computer Science\\
	    University of Texas at Austin\\
	    {\tt mooney@cs.utexas.edu}}

\date{}

\begin{document}

\maketitle

\begin{abstract}
We present results on combining supervised and unsupervised methods to ensemble multiple systems for two popular Knowledge Base Population (KBP) tasks, Cold Start Slot Filling (CSSF) and Tri-lingual Entity Discovery and Linking (TEDL). We demonstrate that our combined system along with auxiliary features outperforms the best performing system for both tasks in the 2015 competition, several ensembling baselines, as well as the state-of-the-art stacking approach to ensembling KBP systems. The success of our technique on two different and challenging problems demonstrates the power and generality of our combined approach to ensembling. 
\end{abstract}

\section{Introduction}
Using {\it ensembles} of multiple systems is a standard approach to improving
accuracy in machine learning \cite{dietterich:bkchapter00}. Ensembles have been
applied to a wide variety of problems in natural language processing, including
parsing \cite{henderson:emnlp99}, word sense disambiguation
\cite{pedersen:naacl00}, sentiment analysis
\cite{whitehead:bkchapter10} and information extraction (IE) \cite{florian:acl03,mcclosky:biomed12}.
Recently, using {\it stacking} \cite{wolpert:nn92} to ensemble 
IE systems was shown to give state-of-the-art results on
slot-filling for {\it Knowledge Base Population} (KBP)
\cite{viswanathan:acl15}.  Stacking uses supervised learning to train a meta-classifier to combine multiple system outputs; therefore, it requires historical data on the performance of each system on a
corpus of labeled training data. Viswanathan et al.\  \shortcite{viswanathan:acl15}
use data from the 2013 KBP slot-filling competition for training and 
then test on the data from the 2014 competition, therefore they can only
ensemble the {\it shared systems} that participated in both years. 

However, in some situations, we would like to ensemble systems for
which we have no historical performance data. For example, due to privacy, some companies or agencies may
not be willing to share their raw data but their final models or meta-level model output. Simple methods such as
voting permit ``unsupervised'' ensembling, and several more
sophisticated methods have also been developed for this
scenario \cite{wang:tac13}.  However, such methods fail to exploit
supervision for those systems for which we {\it do} have
training data.  Therefore, we present an approach that utilizes
supervised {\it and} unsupervised ensembling to exploit the advantages
of both.  We first use unsupervised ensembling to combine systems
without training data, and then use stacking to combine this ensembled
system with other systems with available training data.

Using this new approach, we demonstrate new state-of-the-art results on two separate tasks in the well-known NIST KBP challenge -- {\it Cold Start Slot-Filling} (CSSF)\footnote{\label{1}http://www.nist.gov/tac/2015/KBP/ColdStart/guidelines.html} and the {\it Tri-lingual Entity Discovery and Linking} (TEDL) \cite{ji:tac15}. Our approach outperforms the best individual system as well as other ensembling methods  (such as stacking only the shared systems) on both tasks in the most recent 2015 competition; verifying the generality and power of combining supervised and unsupervised ensembling. There was been some work in the past on combining multiple supervised and unsupervised models using graph-based consensus maximization \cite{gao:nips09}, however we show that it does not do as well as our stacking method. We also propose two new auxiliary features for the CSSF task and verify that incorporating them in the combined approach improves performance.

\section{Background}
For the past several years, NIST has conducted
the English Slot Filling (ESF) and the Entity Discovery and Linking (EDL) tasks in the Knowledge Base Population (KBP) track as a part of the Text Analysis Conference (TAC). In 2015, the ESF task \cite{surdeanu:tac13,surdeanu:tac14} was replaced by the Cold Start Slot Filling (CSSF) task\footnote{http://www.nist.gov/tac/2015/KBP/ColdStart/index.html} which requires filling specific slots of information for a given set of query entities  based on a supplied text corpus. For the 2015 EDL task, two new foreign languages were introduced -- Spanish and Chinese as well as English -- and thus the task was renamed  Tri-lingual Entity Discovery and Linking (TEDL) \cite{ji:tac15}. The goal was to discover entities for all three languages based on a supplied text corpus as well as link these entities to an existing English Knowledge Base (KB) or cluster the mention with a NIL ID if the system could not successfully link the mention to any entity in the KB.

For CSSF, the participating systems
employ a variety of techniques such as such as relevant document extraction, relation-modeling, open-IE and inference \cite{finin:tac15,soderland:tac15,kisiel:tac15}. The top performing 2015 CSSF system \cite{angeli:tac15} leverages both distant
supervision \cite{mintz:acl09} and pattern-based relation extraction. Another system, {\it UMass\_IESL} \cite{roth:tac15}, used distant supervision, rule-based extractors, and semisupervised matrix embedding methods. The top performing 2015 TEDL system used a combination of deep neural networks and CRFs for mention detection and a language-independent probabilistic disambiguation model for entity linking  \cite{sil:tac15}. 

Given the diverse CSSF and TEDL systems available, it is
productive to  ensemble them, which has been shown to improve performance on slot filling \cite{viswanathan:acl15}. However, their stacking method relies on past training data and thus cannot be used for systems that did not participate in previous years. On the other hand, constrained optimization techniques, that do not crucially rely on past data to aggregate confidence scores across multiple systems for the slot-filling task have also been explored \cite{wang:tac13}. However, there has been no past work on ensembling for the TEDL task.

{\it Stacking} \cite{sigletos:jmlr05,wolpert:nn92} has not been used to combine supervised and unsupervised methods for ensembling KBP systems. In this paper, we introduce the novel idea of combining the supervised stacking approach with the unsupervised constrained optimization approach, improving performance on two different KBP tasks.

\section{Overview of KBP Tasks}
In this section we give a short overview of each of the KBP tasks considered in this paper.

\subsection{Cold Start Slot Filling}
The goal of CSSF
is to collect information (fills) about specific attributes (slots) for a set of entities (queries) from a given corpus. The queries entities can be a person (PER), organization (ORG) or geo-political entity (GPE). The slots are fixed and the 2015 task also included the inverse of each slot, for example the slot org:subsidiaries and its inverse org:parents. Some slots (like per:age) are
{\it single-valued} while others (like per:children) are {\it list-valued} i.e., they
can take multiple slot fillers. 

The input for  CSSF is a set of queries and the corpus in which to look for information. The queries are provided in an XML format that includes an ID for the query, the name of the entity, and the type of entity (PER, ORG or GPE). The corpus consists of documents in XML format from discussion forums, newswire and the Internet,  each identified by a unique ID. The output is a set of slot fills for each query. Along with the slot-fills, systems must also provide its {\it provenance} in the corpus in the form {\it docid:startoffset-endoffset}, where {\it docid} specifies a source
document and the offsets demarcate the text in this document containing the
extracted filler. Systems also provide a confidence score to indicate their certainty in the extracted information.

\subsection{Tri-lingual Entity Discovery and Linking}
The goal of TEDL is to discover all entity mentions in a corpus with English, Spanish and Chinese documents. The entities can be a person (PER),  organization (ORG), geo-political entity (GPE),  facility (FAC), or location (LOC). The FAC and LOC entity types were newly introduced in 2015. The extracted mentions are then linked to an existing English KB entity using its ID. If there is no KB entry for an entity, systems are expected to cluster all the mentions for that entity using a NIL ID.  

The input is a corpus of documents in the three languages and an English KB of entities, each with a name,  ID,  type, and several relation tuples that allow systems to disambiguate entities. The output is a set of extracted mentions,  each with a string, its provenance in the corpus, and a corresponding KB ID if the system could successfully link the mention, or else a mention cluster with a NIL ID. Systems can also provide a confidence score for each mention.

\section{Algorithm}
\begin{figure}
\centering
\includegraphics[height=6.2cm]{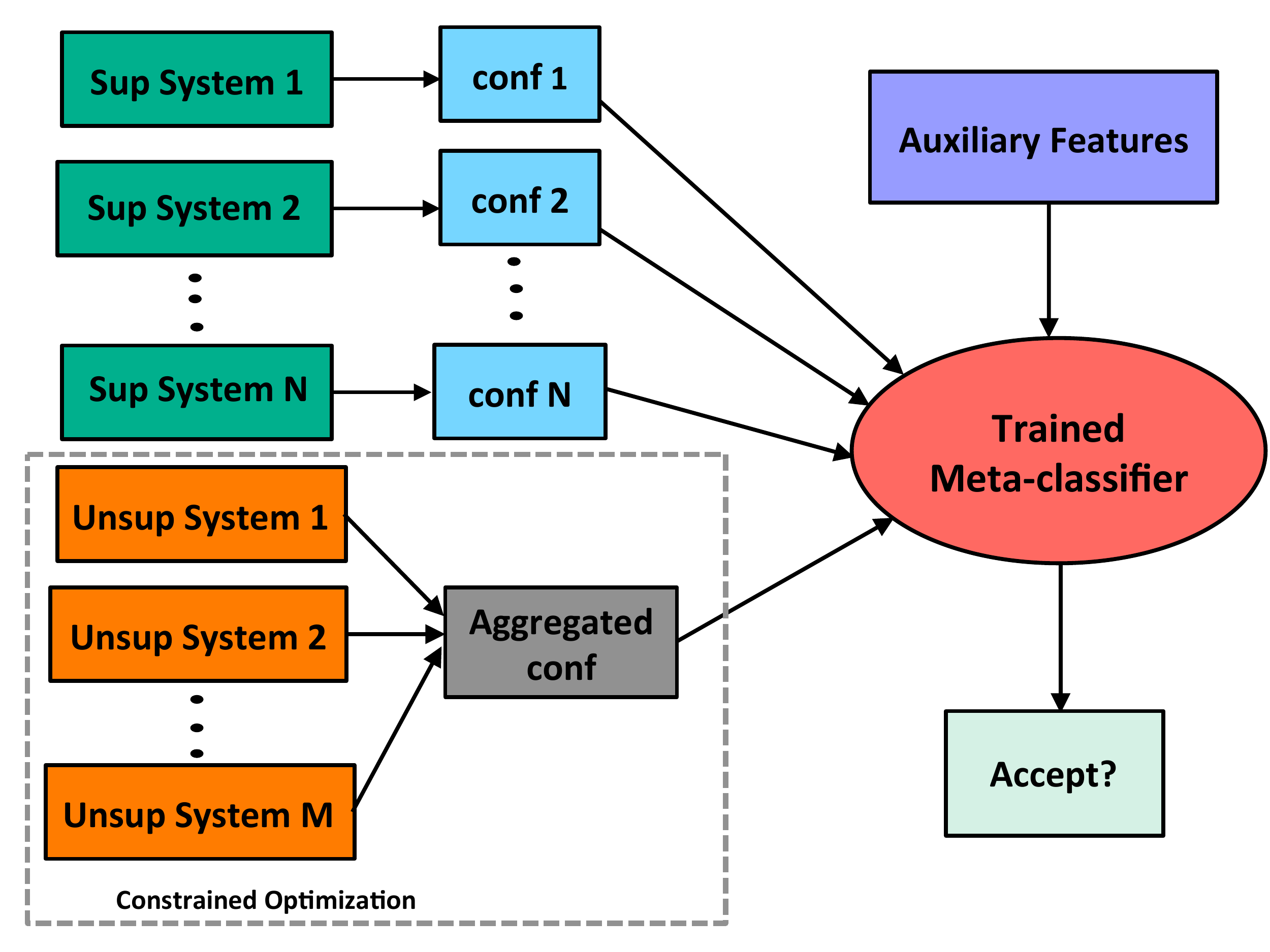}
\caption{Our stacking approach to combining supervised and unsupervised ensembles for KBP.}
\label{fig:supunsup}
\end{figure}

This section describes our approach to ensembling both supervised and unsupervised methods. Figure~\ref{fig:supunsup} shows an overview of our system which trains a final meta-classifier for combining multiple systems using confidence scores and other auxiliary features depending on the task. 

\subsection{Supervised Ensembling Approach}
For the KBP systems that are common between years, we have training data for supervised learning. We use the stacking method described in \newcite{viswanathan:acl15} for these shared systems. The idea is to combine
 predictions by training a ``meta-classifier'' to weight and combine multiple models using their confidence scores as features.  By
training on a set of supervised data that is disjoint from that used to train the individual models, it learns how to combine their results into an improved ensemble model. The output of the ensembling system is similar to the output
of an individual system, but it productively aggregates results from different systems. In a final {\it post-processing} step, the outputs that get classified as ``correct" by the classifier are kept while the others are removed from the output.

The meta-classifier makes a binary decision for each distinct output represented as a {\it key-value} pair.
For the CSSF task, the {\it key} for ensembling multiple systems is a query along with a slot type, for example, per:age of ``Barack Obama'' and the {\it value} is a computed {\it slot fill}. For the TEDL task, we define the key to be the {\it KB (or NIL) ID} and the value to be a {\it mention}, that is a specific reference to an entity in the text.   In Figure~\ref{fig:supunsup}, the top half shows these supervised systems for which we have past training data.

\subsection{Unsupervised Ensembling Approach}
\label{sec:unsupervised}
Only $38$ of the $70$ systems that participated in CSSF 2015 also participated in 2014, and only $24$ of the $34$ systems that participated in TEDL 2015 also participated in the 2014 EDL task. Therefore, many KBP systems in 2015 were new and did not have past training data needed for the supervised approach. In fact, some of the new systems performed better than the shared systems, for example the {\it hltcoe} system did not participate in $2014$ but was ranked $4^{th}$ in the 2015 TEDL task \cite{ji:tac15}.  We first ensemble these unsupervised systems using the constrained optimization approach described by \newcite{wang:tac13}. Their approach is specific to the English slot-filling task and also relies a bit on past data for identifying certain parameter values. Below we describe our modifications to their approach so that it can be applied to both KBP tasks in a purely unsupervised manner. The bottom half of  Figure~\ref{fig:supunsup} shows the ensembling of the systems without historical training data.

The approach in \newcite{wang:tac13} aggregates the “raw” confidence values produced by individual KBP systems to arrive at a single aggregated confidence value for each key-value. Suppose that $V_1, \dots , V_M$ are the $M$ distinct values produced by the systems and $N_i$ is the number of times the value $V_i$ is produced by the systems. Then \newcite{wang:tac13} produce an aggregated confidence by solving the following optimization problem:

\begin{equation}
\label{eq:min}
\begin{split}
\min_{0\leq x_i\leq 1}\sum_{i=1}^{M}\sum_{j=1}^{N_i}w_{ij}\left(x_i-c_i\left(j\right)\right)^2 , \\
\end{split}
\end{equation}
where $c_i$ denotes the raw confidence score and $x_i$ denotes the aggregated confidence score for $V_i$, $w_{ij} \geq 0$ is a non-negative weight assigned to each instance. Equation~\ref{eq:min} ensures that the aggregated confidence score is close to the raw score as well as proportional to the agreement among systems on a value for a given key. Thus for a given key, if a system's value is also produced by multiple other systems, it would have a higher score than if it were not produced by any other system. The authors use the inverse ranking of the average precision previously achieved by individual systems as the weights in the above equation. However since we use this approach for systems that we do not have historical data, we use uniform weights across all unsupervised systems for both the tasks. 

\begin{figure*}[!ht]
\centering
\begin{tabular}{cc}
\includegraphics[width=80mm]{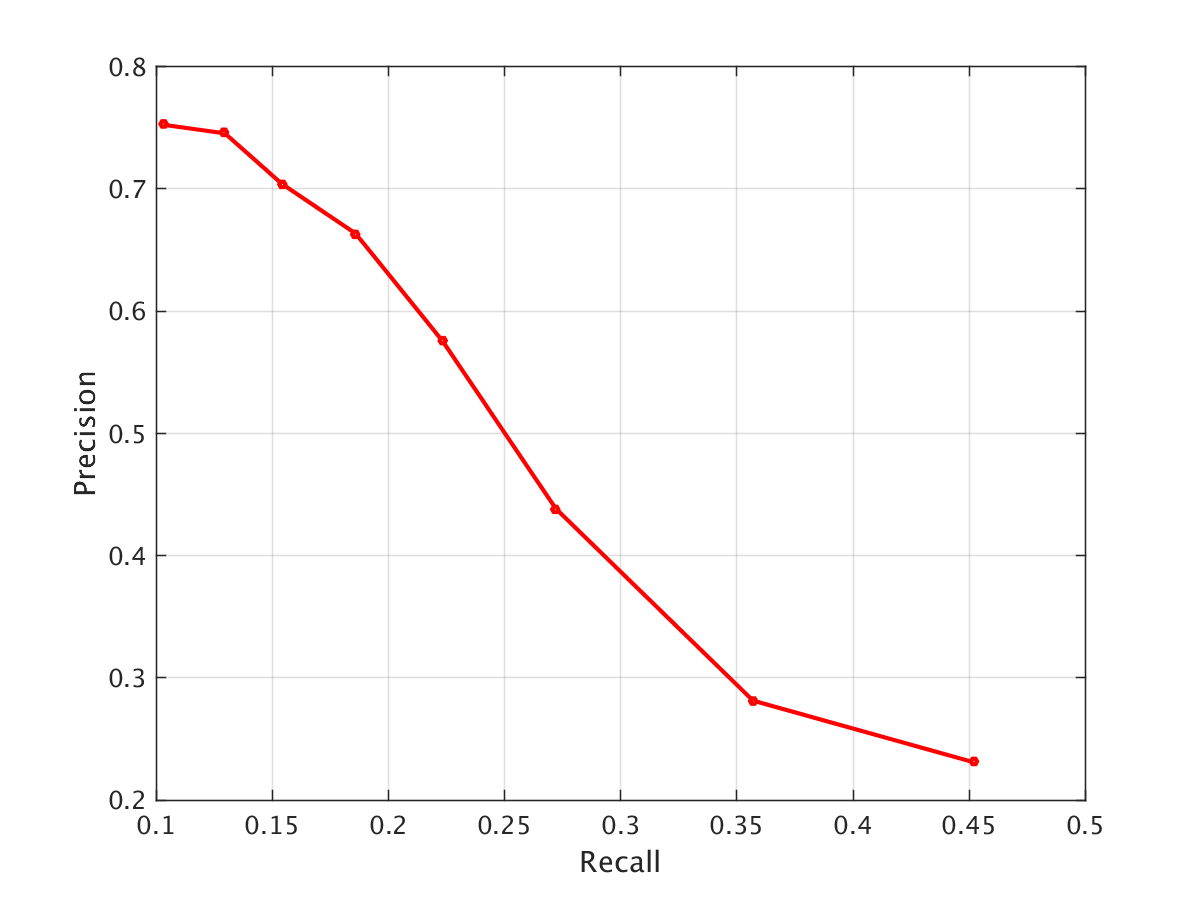} & \includegraphics[width=80mm]{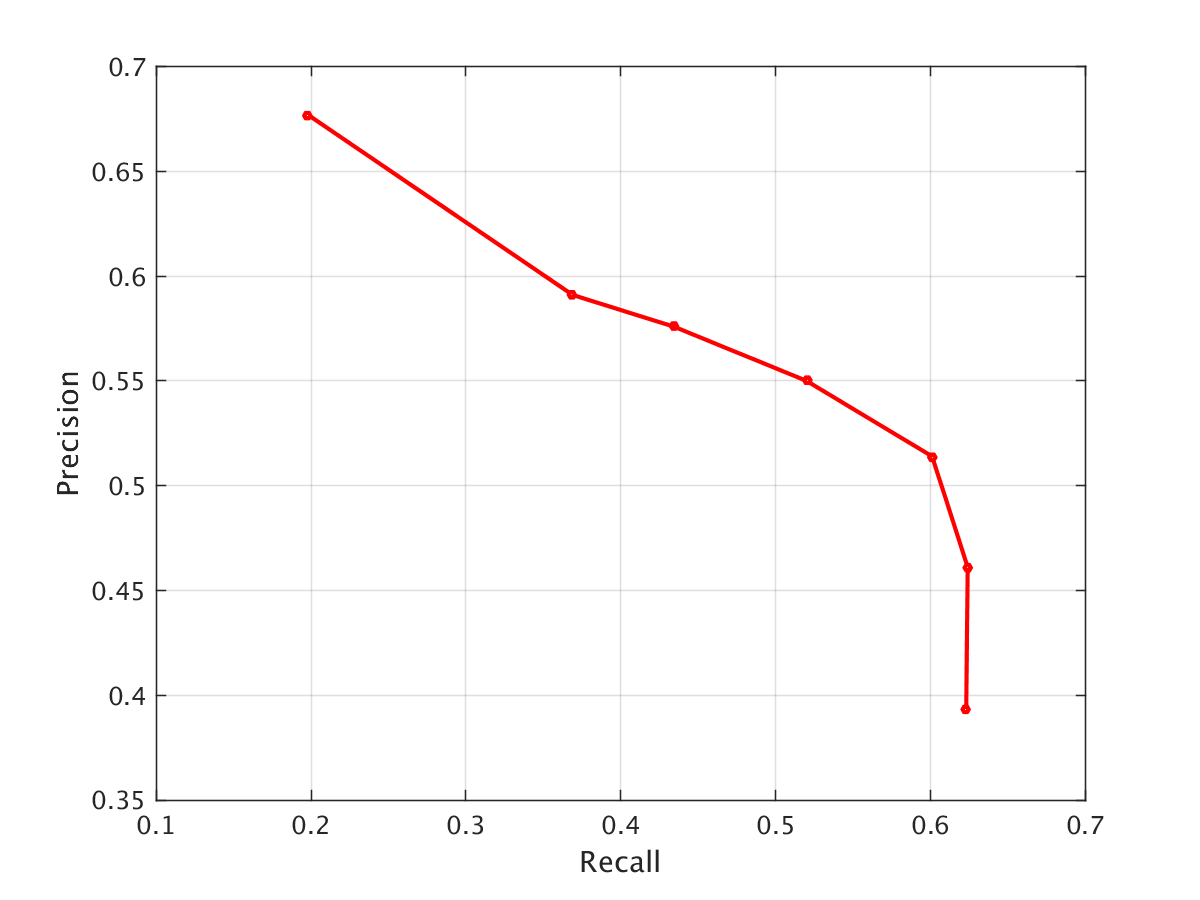}\\
CSSF systems & TEDL systems\\
\end{tabular}
\caption{Precision-Recall curves for identifying the best voting performance on the two KBP tasks}
\label{fig:pr}
\end{figure*}

Equation~\ref{eq:min} is subject to certain constraints on the confidence values depending on the task. For the slot-filling task, the authors define two different constraints based on whether the slot type is single valued or list valued. For single-valued slot types, only one slot value can be correct and thus the constraint is based on the mutual exclusion property of the slot values:
\begin{equation}
\label{eq:cons}
P(V_1) + P(V_2) + \cdots + P(V_M) \leq 1
\end{equation}
This constraint allows only one of the slot values to have a substantially higher probability compared to rest. On the other hand, for list-valued slot types, the $1$ in the above equation is replaced by the value $n_c/n$ where $n_c$ is the average number of correct slot fills for that slot type across all entities in the previous year and $n$ is the total number of slot fills for that slot type across all entities. This approach to estimating the number of correct values can be thought of as {\it collective precision} for the slot type achieved by the set of systems. For the newly introduced slot inverses in 2015, we use the same ratio as that of the corresponding original slot type. Thus the slot type {\it per:parents} (new slot type) would have the same ratio as that of {\it per:children}.

For the TEDL task, we use the KB ID as the key and thus use the entity type for defining the constraint on the values. For each of the entity types (PER, ORG and GPE) we replace the quantity on the right hand side in Equation~\ref{eq:cons} by the ratio of the average number of correct values for that entity type in 2014 to the total number of values for that entity type, across all entities. For the two new entity types instroduced in 2015 (FAC and LOC), we use the same ratio as that of GPE because of their semantic similarities.
 
The output from this approach for both tasks is a set of key-values with aggregated confidence scores across all unsupervised systems which go directly into the stacker as shown in Figure~\ref{fig:supunsup}. Using the aggregation approach as opposed to directly using the raw confidence scores allows the classifier to meaningfully compare confidence scores across multiple systems although they are produced by very diverse systems.  

Another unsupervised ensembling we experimented with in place of the constrained optimization approach is the Bipartite Graph based Consensus Maximization (BGCM) approach by \newcite{gao:nips09}. The authors introduce BGCM as a way of combining supervised and unsupervised models for a given task. So we use their approach for the KBP tasks and compare it to our stacking approach to combining supervised and unsupervised systems, as well as an alternative approach to ensembling the unsupervised systems before feeding their output to the stacker. The idea behind BGCM is to cast the ensembling task as an optimization problem on a bipartite graph, where the objective function favors the smoothness of the prediction over the graph, as well as penalizing deviations from the initial labeling
provided by supervised models. The authors propose to consolidate a classification solution by maximizing the consensus among both supervised predictions and unsupervised constraints. They show that their algorithm outperforms the component models on ten out of eleven classification tasks across three different datasets.

\subsection{Combining the Supervised and Unsupervised Approaches}
Our new approach combines these supervised and unsupervised approaches using a stacked meta-classifier as the final arbiter for accepting a given key-value. Most KBP teams submit multiple variations of their system. Before running the supervised and unsupervised approaches discussed above, we first combine multiple runs of the same team into one. Of the $38$ CSSF systems from $10$ teams for which we have 2014 data for training and the $32$ systems from $13$ teams that do not have training data, we combine the runs of each team into one to ensure diversity of the final ensemble (since different runs from the same team tend to be minor variations). For the slot fills that were common between the runs of a given team, we compute an average confidence value, and then add any additional fills that are not common between runs. Thus, we obtained $10$ systems (one for each team) for which we have supervised data for training stacking. Similarly, we combine the $24$ TEDL systems from $6$ teams that have 2014 training data and $10$ systems from $4$ teams that did not have training data into one per team. Thus using the notation in  Figure~\ref{fig:supunsup}, for TEDL, $N=6$ and $M=4$ while for CSSF, $N=10$ and $M=13$.

The output of the unsupervised method produces aggregated confidence scores calibrated across all of the component systems and goes directly into our final meta-classifier. We treat this combination as  a single system which we call  the {\it unsupervised ensemble}. In other words, in order to combine systems that have training data with those that do not, we add the unsupervised ensemble as an additional system to the stacker, thus giving us a total of $N+1$, that is $11$ CSSF and $7$  TEDL systems. Once we have extracted the auxiliary features for each of the $N$ supervised systems and the unsupervised ensemble for both years, we train the stacker on 2014 systems, and test on the 2015 systems.  The unsupervised ensemble for each year is composed of different systems, but hopefully the stacker learns to combine a generic unsupervised ensemble with the supervised systems that are shared across years. This allows the stacker to be the final arbitrator on the correctness of a key-value pair, combining new systems for which we have no historical data with additional systems for which training data {\it is} available. We employ a single
classifier to train and test the meta-classifier using an L1-regularized SVM with a linear kernel \cite{fan:ref08} (other classifiers gave
similar results).

\begin{table*}[!ht]
\centering
\begin{tabular}{| c | c | c | c | }
\hline
\textbf{Methodology} & \textbf{Precision} & \textbf{Recall} & \textbf{F1} \\
 \hline
 \hline
Combined stacking and constrained optimization with auxiliary features&0.4679&\textbf{0.4314}&\textbf{0.4489}\\
Top ranked SFV system in 2015 \cite{rodriguez:tac15}&0.5101&0.1812&0.4228\\
Combined stacking and constrained optimization without new features& 0.4789&0.3588&0.4103\\
Stacking using BGCM instead of constrained optimization&\textbf{0.5901} &0.3021&0.3996\\
BGCM for combining supervised and unsupervised outputs&0.4902&0.3363&0.3989\\
Stacking approach described in \cite{viswanathan:acl15}& 0.5084&0.2855&0.3657\\
Top ranked CSSF system in 2015 \cite{angeli:tac15}&  0.3989&0.3058&0.3462\\
Oracle Voting baseline (3 or more systems must agree)&0.4384&0.2720&0.3357\\
Constrained optimization approach described in \cite{wang:tac13} & 0.1712& 0.3998&0.2397\\ \hline
\end{tabular}
\caption{Results on 2015 Cold Start Slot Filling (CSSF) task using the official NIST scorer}
\label{table:results1}
\end{table*}

\subsection{Auxiliary Features for Stacking}
Along with the confidence scores, we also include auxiliary features which provide additional context for improving the meta-classifier  \cite{viswanathan:acl15}. For CSSF, the slot type (e.g. per:age) is used as an auxiliary feature, and in TEDL, we use the entity type. For slot-filling, features related to the provenance of the fill have also been used \cite{viswanathan:acl15}. We include these for CSSF, along with two new features.

The two novel features measure similarity between specific documents. The 2015 CSSF task had a much smaller corpus of shorter documents compared to the previous year's slot-filling corpus \cite{ellis:tac15,surdeanu:tac14}. Thus, the provenance feature of \newcite{viswanathan:acl15} did not sufficiently capture the reliability of a slot fill based on where it was extracted. Slot filling queries were provided to participants in an XML format that included the query entity's ID, name, entity type, the document where the entity appears, and beginning and end offsets in the document where that entity appears. This allowed the participants to disambiguate query entities that could potentially have the same name but refer to different entities. Below is a sample query from the 2015 task:
\lstset{language=XML}
\begin{lstlisting}[basicstyle=\footnotesize]
  <query id="CSSF15_ENG_0006e06ebf">
    <name>Walmart</name>
    <docid>ad4358e0c4c18e472c13bbc27a6b7ca5</docid>
    <beg>232</beg>
    <end>238</end>
    <enttype>org</enttype>
    <slot0>org:date_dissolved</slot0>
  </query>
\end{lstlisting}
The \textless docid\textgreater\ tag refers to the document where the query entity appears, which we will call the {\it query document}. 

Our first new feature involves measuring the similarity between this query document and the provenance document that is provided by a given system. We represent the query and provenance documents as standard TF-IDF weighted vectors and use cosine similarity to compare documents. Therefore every system that provides a slot fill, also provides the provenance for the fill and thus has a similarity score with the query document. If a system does not provide a particular slot fill then its document similarity score is simply zero. This feature is intended to measure the degree to which the system's provenance document is referencing the correct query entity.

Our second new feature measures the document similarity between the {\it provenance documents} that different systems provide. Suppose for a given query and slot type, $n$ systems provide the same slot fill. For each of the $n$ systems, we measure the average document cosine similarity between the system's provenance document and those of the other $n-1$ systems. The previous approach simply measured whether systems agreed on the exact provenance document. By softening this to take into account the {\it similarity} of provenance documents, we hope to more flexibly measure provenance agreement between systems.

For TEDL, we only use the entity type as an auxiliary feature and leave the development of more sophisticated features as future research.

\subsection{Post-processing}
Once we obtain the decisions on each of the key-value pairs from the stacker, we perform some final post-processing. For CSSF, this is straight forward. Each list-valued slot fill that is classified as correct is included in the final output. For single-valued slot fills, if they are multiple fills that were classified correctly for the same query and slot type, we include the fill with the highest meta-classifier confidence.

For TEDL, for each entity mention link that is classified as correct, if the link is a KB cluster ID then we include it in the final output, but if the link is a NIL cluster ID then we keep it aside until all mention links are processed. Thereafter, we resolve the NIL IDs across systems since NIL ID's for each system are unique. We merge NIL clusters across systems into one if there is at least one common entity mention among them. Finally, we give a new NIL ID for these newly merged clusters.

\begin{table*}[!ht]
\centering
\begin{tabular}{| c | c | c | c | }
\hline
\textbf{Methodology} & \textbf{Precision} & \textbf{Recall} & \textbf{F1} \\
 \hline
 \hline
Combined stacking and constrained optimization &0.686&\textbf{0.624} &\textbf{0.653}\\
Stacking using BGCM instead of constrained optimization& 0.803 &0.525  &0.635\\
BGCM for combining supervised and unsupervised outputs&0.810&0.517&0.631\\
Stacking approach described in \cite{viswanathan:acl15}& \textbf{0.814}&0.508  &0.625\\ 
Top ranked TEDL system in 2015 \cite{sil:tac15}&  0.693&0.547 &0.611\\
Oracle Voting baseline (4 or more systems must agree)&0.514 &0.601 &0.554\\
Constrained optimization approach & 0.445&0.176 &0.252\\ \hline
\end{tabular}
\caption{Results on 2015 Tri-lingual Entity Discovery and Linking (TEDL) task using the official NIST scorer and the CEAFm metric}
\label{table:results2}
\end{table*}

\section{Experimental Results}
\label{sec:results}

This section describes a comprehensive set of experiments evaluating ensembling for both KBP tasks using the algorithm described in the previous section, comparing our full system to various ablations and prior results. All results were obtained using the official NIST scorers for the tasks provided after the 
competition ended.\footnote{http://www.nist.gov/tac/2015/KBP/ColdStart/tools.html, 
https://github.com/wikilinks/neleval} We compare our results to several baselines. We apply the purely supervised approach of \newcite{viswanathan:acl15} to systems that are common between 2014 and 2015, and also the constrained optimization approach of \newcite{wang:tac13} on all the 2015 KBP systems. We also compare our combined stacking approach to Bipartite Graph based Consensus Maximization (BGCM)  \cite{gao:nips09} in two ways. First, we use BGCM in place of the constrained optimization approach to ensemble unsupervised systems while keeping the rest of our pipeline the same. Secondly, we also compare to combining both supervised and unsupervised systems using BGCM instead of stacking. We also include a voting baseline for ensembling the system outputs. For this approach, we vary the
threshold on the number of systems that must agree to identify an ``oracle'' threshold that
results in the highest F1 score for 2015 by plotting a Precision-Recall curve and finding the best F1 score for the voting baseline on both the KBP tasks. Figure~\ref{fig:pr} shows the plots for finding this ``oracle'' threshold for each of the KBP tasks. At each step we add one more to the number of systems that must agree on a key-value. We find that for CSSF, a threshold of $3$ or more systems and for TEDL a threshold of $4$ or more systems gives us the best resulting F1 for voting.

Tables \ref{table:results1} and \ref{table:results2} show the results for CSSF and TEDL respectively.
Our full system, which combines supervised and unsupervised ensembling performed the best on both tasks.  TAC-KBP also includes the Slot Filler Validation (SFV) task\footnote{http://www.nist.gov/tac/2015/KBP/SFValidation/index.html} where the goal is to ensemble/filter outputs from multiple slot filling systems. The top ranked system in 2015 \cite{rodriguez:tac15}  does substantially better than many of the other ensembling approaches,  but it does not do as well as our best performing system. The purely supervised approach of  \newcite{viswanathan:acl15} performs substantially worse, although still outperforming the top-ranked individual system in the 2015 competition. This approach only uses the common systems from 2014, thus ignoring approximately half of the systems. The approach of \newcite{wang:tac13} performs very poorly by itself; but when combined with stacking gives a boost to recall and thus the overall $F1$. Note that all our combined methods have a substantially higher recall and thus highlighting the importance of the unsupervised ensemble. The oracle voting baseline also performs very poorly indicating that naive ensembling is not advantageous. 

For TEDL, our combined approach  gives the best overall performance, beating the top-ranked system for TEDL 2015. The TEDL evaluation provides three different approaches to measuring Precision, Recall and F1. First is entity discovery, second is entity linking and last is mention CEAF \cite{ji:tac15}. The mention CEAF metric finds the optimal alignment between system and gold standard clusters, and then evaluates precision and recall micro-averaged
over mentions. We obtained similar results on all three evaluations and thus only include the mention CEAF score in this paper. The purely supervised stacking approach over shared systems does not do as well as any of our combined approaches even though it beats the best performing system (i.e. IBM) in the 2015 competition \cite{sil:tac15}.  The relative ranking of the approaches is similar to those obtained on the CSSF task, thus proving that our approach is very general and provides improved performance on two quite different and challenging problems.


\section{Related Work}
Stacking has been previously applied to several problems in NLP such as collective document classification \cite{kou:acl07}, named entity recognition \cite{florian:acl02}, stacked dependency parsing \cite{martins:acl08} and joint Chinese word segmentation
and part-of-speech tagging \cite{sun:acl11}.
Although \newcite{viswanathan:acl15} applied stacking to KBP slot filling, we extend their approach with new auxiliary features and combine supervised {\it and} unsupervised systems for both CSSF {\it and} TEDL.

A fast and scalable collective entity linking method that relies on stacking was proposed by \newcite{he:emnlp13}. They stack a global predictor on top of a local predictor to collect coherence information from neighboring decisions. Biomedical entity extraction using a stacked ensemble of an SVM and CRF was shown to outperform individual components as well as voting baselines \cite{ekbal:elsevier13}.

Stacking for information extraction has been
demonstrated to outperform both majority voting and weighted voting methods \cite{sigletos:jmlr05}. The FAUST system for biomolecular even extraction uses model combination strategies such as voting and stacking and was placed first in three of the four BioNLP tasks in 2011  \cite{riedel:acl11}. Google's Knowledge Vault system \cite{dong:kdd14} combines four
diverse extraction methods by building a boosted decision stump classifier
\cite{reyzin:ml2006}.  For each proposed fact, the classifier considers both the
confidence value of each extractor and the number of responsive documents found
by the extractor. 


\section{Conclusion and Future Work}
This paper has presented experimental results on two diverse KBP tasks, showing that a novel stacking-based approach to ensembling both supervised and unsupervised systems  is  very promising.  The approach provides an overall F1 score of $44.9\%$ on
2015 KBP CSSF task and CEAFm F1 of $65.3\%$ on 2015 KBP TEDL, outperforming the top ranked systems from both 2015 competitions as well as several other baseline ensembling methods,  thereby achieving a new state-of-the-art for both of these important, challenging tasks.
We found that adding the unsupervised ensemble along with the shared systems increased the recall substantially, 
highlighting the importance of  utilizing systems that do not have historical training data.

As discussed in Section~\ref{sec:results}, two new auxiliary stacking features for slot-filling  based on provenance similarity with the query document and document similarity across systems improved CSSF performance substantially.  In the future, we hope to develop similar auxiliary features for EDL. The input for TEDL includes a KB with several relational tuples for each entity. Similarity of this relational information to the context of a mention could be a useful auxiliary feature. Another feature for evaluating the reliability of agreement on a particular mention link could be a measure of how often the same systems agree on other linking decisions. 

\section{Acknowledgment}
This research was supported in part by the DARPA DEFT program under AFRL grant FA8750-13-2-0026 and by MURI ARO grant W911NF-08-1-0242.
\bibliographystyle{acl2016}
\bibliography{acl2016}
\end{document}